\pdfoutput=1
\def\year{2022}\relax
\documentclass[letterpaper]{article} 
\usepackage{aaai22}  
\usepackage{times}  
\usepackage{helvet}  
\usepackage{courier}  
\usepackage[hyphens]{url}  
\usepackage{graphicx} 
\usepackage{natbib}  
\usepackage{caption} 
\DeclareCaptionStyle{ruled}{labelfont=normalfont,labelsep=colon,strut=off} 
\usepackage{amssymb}  
\usepackage{amsmath}  
\usepackage[ruled,linesnumbered]{algorithm2e}  
\usepackage{amsthm}
\usepackage{array}
\usepackage{graphicx}
\usepackage{multirow}
\usepackage{arydshln}
\usepackage{soul}
\usepackage{tabularx}
\usepackage{booktabs}
\usepackage{enumitem}
\usepackage{xcolor}
\usepackage{lipsum}

\newcommand\blfootnote[1]{%
  \begingroup
  \renewcommand\thefootnote{}\footnote{#1}%
  \addtocounter{footnote}{-1}%
  \endgroup
}
\makeatletter
\newcommand{\thickhline}{
\noalign {\ifnum 0=`}\fi \hrule height 0.8pt
\futurelet \reserved@a \@xhline
}

\usepackage{newfloat}
\usepackage{placeins}
\usepackage{listings}
\title{Idiomatic Expression Paraphrasing without Strong Supervision}
 \author {
     Jianing Zhou\textsuperscript{\rm 1},
     Ziheng Zeng\textsuperscript{\rm 1},
     Hongyu Gong\textsuperscript{\rm 1,2}\blfootnote{The work was done while Hongyu Gong was at UIUC\\},
     Suma Bhat\textsuperscript{\rm 1}
 }
 \affiliations {
     \textsuperscript{\rm 1} University of Illinois at Urbana-Champaign\\
     \textsuperscript{\rm 2} Facebook AI\\
     \textsuperscript{\rm 1} \{zjn1746, zzeng13,spbhat2\}@illinois.edu\\ 
     \textsuperscript{\rm 2} hygong@fb.com
 }

\usepackage{bibentry}
\begin{document}

\maketitle
\begin{abstract}
Idiomatic expressions (IEs) play an essential role in natural language. In this paper, we study the task of idiomatic sentence paraphrasing (ISP), which aims to paraphrase a sentence with an IE by replacing the IE with its literal paraphrase. 
The lack of large-scale corpora with idiomatic-literal parallel sentences is a primary challenge for this task, for which we consider two separate solutions. First, we propose an unsupervised approach to ISP, which leverages an IE's contextual information and definition and does not require a parallel sentence training set. Second, we propose a weakly supervised approach using back-translation to jointly perform paraphrasing and generation of sentences with IEs to enlarge the small-scale parallel sentence training dataset.  Other significant derivatives of the study include a model that replaces a literal phrase in a sentence with an IE to generate an idiomatic expression and a large scale parallel dataset with idiomatic/literal sentence pairs. 
The effectiveness of the proposed solutions compared to competitive baselines is seen in the relative gains of over 5.16 points in BLEU, over 8.75 points in METEOR, and over 19.57 points in SARI when the generated sentences are empirically validated on a parallel dataset using automatic and manual evaluations. 
We demonstrate the practical utility of ISP as a preprocessing step  
in En-De machine translation.
\end{abstract}
\section{Introduction} \label{sec:introduction}
Idiomatic expressions (IEs) are multi-word expressions  whose meaning cannot be inferred from that of their constituent words, a property known as non-compositionality \cite{nunberg1994idioms}. These expressions have varied forms, ranging from fixed expressions such as \textit{by the way} to figurative constructions such as \textit{born with a silver spoon in one's mouth}.  Not only are IEs  an essential component of a native speakers'  lexicon \cite{jackendoff1995boundaries}, they also render language more  natural  \cite{sprenger2003fixed}.  Their non-compositionality has been the classical `pain in the neck' for NLP applications \cite{sag2002multiword,salton2014empirical}
and studies to make these applications idiom-aware, either by identifying them before or during the task \cite{nivre2004multiword,nasr2015joint} suggest that IE paraphrasing as a preprocessing step holds promise for NLP. Despite this, research on IE paraphrasing  remains largely under-explored \cite{zhou2021solving}.  While most IE processing studies have focused on their identification and detection \cite{gong2017geometry, liu2018heuristically, biddle2020leveraging}, 
in this paper, we study the task  of idiomatic sentence paraphrasing (ISP), i.e., automatically paraphrasing IEs into literal expressions.  We refer to a sentence with an IE as an \textit{idiomatic sentence} and to its corresponding sentence where the IE is replaced with a literal phrase as the \textit{literal sentence}.
Table~\ref{tab:dataset} shows examples of idiomatic and literal sentences between which we expect to paraphrase. 
 {Ideally, an ISP system would have an IE span detection stage to detect the presence and span of IEs  \cite{zeng2021idiomatic} and feeds only idiomatic sentences to ISP. Here we study the ISP task on its own and assume the input sentence is idiomatic and the IE span is available. }


\begin{table}[t]
\small
\setlength{\belowcaptionskip}{0.1pt}
    \centering
    \begin{tabular}{m{3.6cm}|m{3.8cm}}
    \thickhline
        \textbf{Idiomatic sentences} & \textbf{Literal sentences}  \\
    \thickhline
        Nature conservation \textbf{\color{red}{runs against the grain}} of current political doctrine. & Nature conservation \textbf{\color{blue}{is contrary to}} current political doctrine. \\
        \hline
        Putting him \textbf{\color{red}{behind bars}} won't serve any purpose, will it? & Putting him \textbf{\color{blue}{in prison}} won't serve any purpose, will it? \\
    \thickhline
    \end{tabular}
    \caption{Examples of idiomatic sentences and corresponding literal sentences. Idioms are highlighted in \textbf{\color{red}{bold red}}, and literal paraphrases are in \textbf{\textcolor{blue}{bold blue}}.}
    \label{tab:dataset}
\end{table}


Semantic simplification using ISP can be used to many ends, including    for making reading more inclusive for populations that struggle to comprehend figurative expressions in everyday text  (e.g., children with the autistic spectrum disorder   \cite{norbury2004factors}). Based on prior studies  \cite{nivre2004multiword, nasr2015joint},  it  could also serve as a  preprocessing step  for downstream  applications---an aspect we explore in this study.   

Successful ISP involves overcoming at least two challenges:  
 (1) The linguistic challenge of handling semantic ambiguity, i.e., ensuring that the  meaning of the IE and that of the literal phrase    match when an IE is polysemous, e.g. the idiom \textit{give her a hand} can  mean both ``applaud her'' and  ``help her,'' and (2) the related resource-challenge of the lack of large-scale parallel literal and idiomatic expressions for training, because a small training set  leads to  the input being unchanged at the output 
 \cite{zhou2021solving}.

Addressing the second challenge is the main focus of this study, whose contributions are summarized below.


\noindent 1. Given the paucity of large-scale parallel datasets of idiomatic-literal sentence pairs, we study  ISP in two machine learning settings. The first is  \textit{unsupervised}, where we consider a zero-resource scenario with neither access to a parallel dataset nor to a lexicon of IEs \textit{during training}, and the second is \textit{weakly-supervised,} where we consider a low-resource scenario with access to a limited but high quality parallel dataset and a large corpus of idiomatic sentences. Our training strategy relies on a back-translation-based augmentation that yields a large parallel dataset. \\
\noindent 2. Compared to competitive supervised baselines the proposed weakly-supervised method shows  performance gains of over 5.16 points in BLEU and over 19.57 points in SARI (automatic evaluation) and superior generation quality (manual evaluation). Despite the lack of supervision, the unsupervised method's performance compares favorably to that of the supervised baselines.\\
{\noindent 3. Our weakly-supervised method yields 
a large  parallel dataset of idiomatic sentences and their  literal counterparts with 1,169 IEs and their 15,627 sentence pairs, which we share for future research.}\footnote{The code and dataset are available at https://github.com/zhjjn/ISP.git.} \\ 
\noindent 4. We demonstrate  the gains to machine translation  only using   ISP as a pre-processing step  via an English-German challenge set \cite{fadaee2018examining}; translating idiomatic sentences after  paraphrasing them to their literal counterparts yielded a  gain of 0.6 points in BLEU.

\section{Related Work} \label{sec:related_work}
ISP was explored as idiomatic expression substitution in \citet{liu-hwa-2016-phrasal} using  a set of pre-defined heuristic rules to extract portions of the idiom's definitions to replace the IE and  then applying various post-processing steps to render the sentence. 
Going beyond this study, ISP relates to three distinct streams of text generation tasks: \textit{paraphrasing},  \textit{style transfer} and \textit{ IE processing}.  

\noindent \textbf{Paraphrasing} 
is to rewrite a given sentence while preserving its original meaning; prior studies include several sequence-to-Sequence (Seq2Seq) models \cite{gupta2018deep} and other controlled generation methods via template \cite{gu2019extract}, syntactic structures \cite{huang2021generating}, or versatile control codes \cite{keskarCTRL2019}. 
Unlike paraphrasing, which is unconstrained, ISP is more stylistically constrained given the paraphrasing of an IE to its literal meaning.

\noindent \textbf{Style Transfer} 
rewrites sentences into those that conform to a target style. This has been studied as distinctive lexical patterns  and syntactic constructions by \citet{krishna-etal-2020-reformulating}, and  as sentiment, formality or authorship manipulation \cite{jhamtani2017shakespearizing,gong2019reinforcement}. 
Our study is different from these prior methods, including the supervised \cite{li2018delete, sudhakar2019transforming} and unsupervised ones \cite{gong2019reinforcement, zeng2020style}, in that our task retains a large portion of the input sentence in the transferred sentence. Besides, we consider a heretofore unexplored nuanced stylist element that is marked by  figurative and non-literal phrases.

{\noindent \textbf{IE processing} tasks consider idiom type classification and idiom token classification \cite{liu2019toward}: idiom type classification \cite{cordeiro2016predicting} determines if a phrase could be used as an IE; and idiom token classification \cite{liu2017representations, liu2019generalized} disambiguates if a given potentially idiomatic expression is used literally or idiomatically in a given context (sentence). Most prior works require the knowledge of the IE \cite{ liu2017representations,liu2019generalized} 
but  recent efforts on idiom span detection \cite{zeng2021idiomatic} have removed the need for IEs'  identity. Our study is in line with the traditional set-up where the IE positions are assumed to be known. }



\section{The Unsupervised Approach} \label{sec:unsupervised_method}
For the zero-resource ISP scenario where no parallel datasets are available during training,  we train a masked conditional sentence generation model  such that given a sentence with a masked word, the model fills the mask using the masked word's definition and part-of-speech (POS) tag. 
The word's definition and POS tags as inputs account for the semantic and the syntactic properties of the filled word. 
During inference, we mask the IE in the sentence to perform ISP while providing the definition of the IE\footnote{A dictionary for accessing the IE definitions is available to the model during inference; the users only provide the sentences.}.
The definitions of the masked word (or the IE during inference) and its POS tag are  available from linguistic resources such as dictionaries and POS taggers.
 Our model, denoted as \textsc{Bart-Ucd}, is unsupervised because its training does not rely on knowing the IEs  nor the direct supervision from a parallel dataset.

Although conceptually similar to \citet{liu-hwa-2016-phrasal}'s setup, \textsc{Bart-Ucd} (1) does not modify or operate on the definitions using pre-determined  dictionary-specific rules; (2) inserts phrases based on the context instead of inserting a fixed chunk from the definition; (3) is naturally applicable to words and IEs with multiple definitions; 
and (4) generates fluent and grammatically correct sentences without burdensome post-processing steps. We exclude the unsupervised method of \citet{liu-hwa-2016-phrasal} as a baseline in our experiments owing to its unavailability and poor replicability. 


\subsection*{Model Architecture} 
\begin{figure}
  \includegraphics[width=\linewidth]{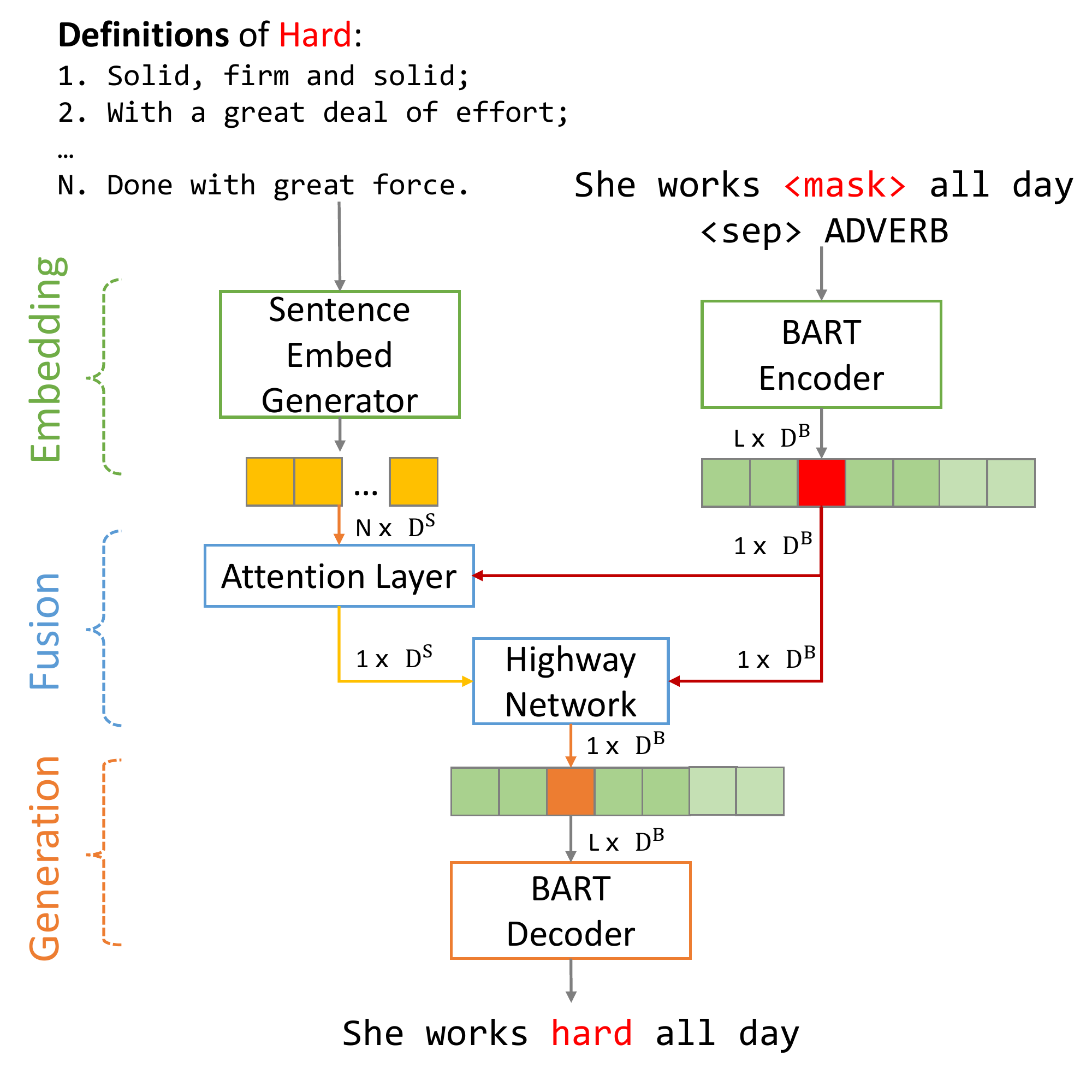}
  \caption{An overview of the unsupervised method. In this example of a \textit{training} instance, the input sentence has a masked word ``hard''. The model takes as input the sentence, the definitions of the word ``hard'' and its POS tag ``ADVERB'' and generates a sentence with the mask filled. 
  }
  \label{fig:unsupervised_model}
\end{figure}

The overall architecture of  \textsc{Bart-Ucd} is illustrated in Figure~\ref{fig:unsupervised_model} and it consists of three stages: (1) the \textit{embedding stage}, (2) the \textit{fusion stage}, and (3) the \textit{generation stage}. In this section, we describe each stage in detail. 

\begin{figure*}
  \includegraphics[width=\linewidth]{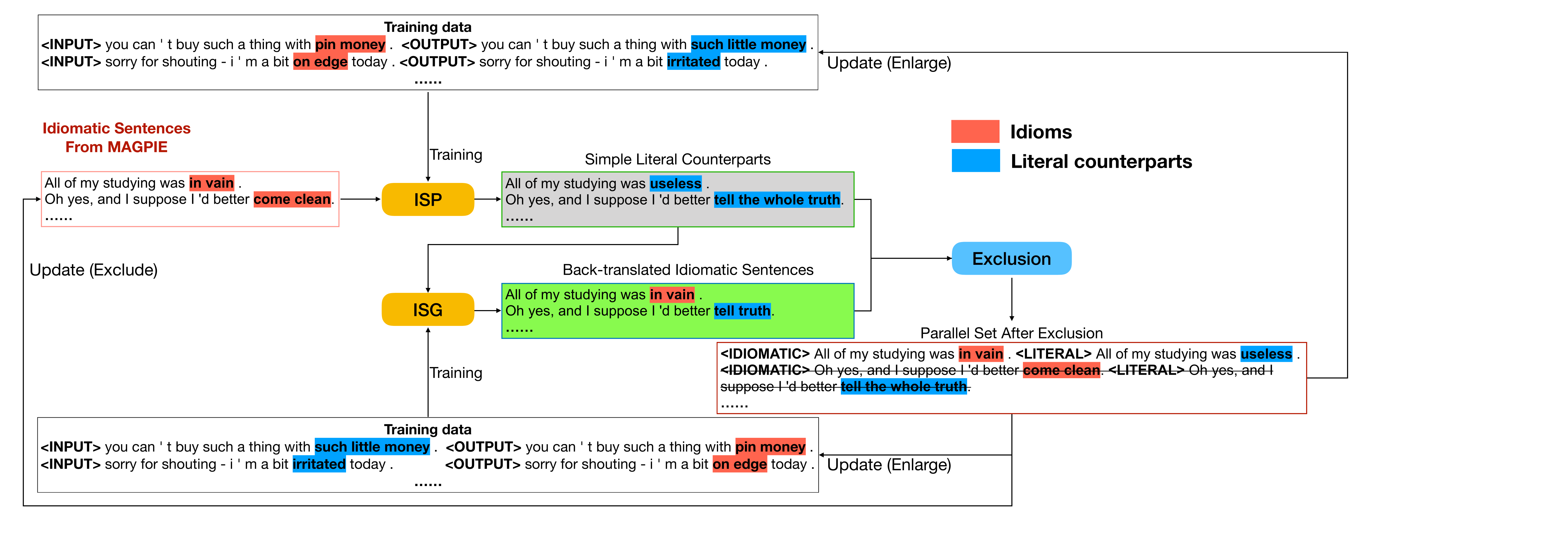}
  \caption{The overview of the weakly supervised method. In each iteration, the method (1) uses the parallel dataset to train an ISP and an ISP  model; (2) constructs augmented parallel pairs; (3) enlarges the parallel dataset with the augmented pairs.}
  \label{fig:supervised_model}
\end{figure*}

\noindent \textbf{The Embedding stage.} 
This stage generates the contextualized word-  and sentence embeddings for the definitions. Specifically, given $[I, \langle\texttt{sep}\rangle, t]$, where $I$ is the masked sentence and $t$ is the POS tag, the model uses a pre-trained BART \cite{liu2020multilingual} encoder to produce contextualized word embeddings $E^{I} \in  \mathbb{R}^{(L+2) \times D^{B}}$, where $|I| = L$.
Then, given a list of $N$ definitions for masked word, the model employs a pre-trained RoBERTa-based \cite{DBLP:journals/corr/abs-1907-11692} sentence embedding generator to generate definition sentence embeddings $E^{D} \in  \mathbb{R}^{N \times D^{S}}$. 
During training, both the BART encoder and the sentence embedding generator are pre-trained and frozen. 

\noindent \textbf{The Fusion stage.} 
This stage combines the definition embeddings $E^{D}$ and the word embedding $E^{B}_{w}$ for the masked token $i_{w}$ and replaces $E^{B}_{w}$ with the combined embedding. Specifically, the model first transforms $E^{D}$ into a single vector $\hat{E}^{D} \in \mathbb{R}^{1 \times D^{S}}$ using an attention mechanism \cite{luong-etal-2015-effective} with $E^{B}_{w}$ as the query to generate the attention weights. Then, the model fuses $\hat{E}^{D}$ and $E^{B}_{w}$ using a highway network \cite{NIPS2015_215a71a1} followed by a linear layer to produce the definition-aware contextualized embedding for $w$, $\tilde{E}^{D}_{w} \in \mathbb{R}^{1 \times D^{B}}$. Based on an empirical observation of improved performance, we replace the original linear + $tanh$ part of the attention mechanism with the highway network. Finally, the model replaces $E^{B}_{w}$ from $E^{B}$ with $\tilde{E}^{D}_{w}$ to produce $\tilde{E}^{D}$ . 

\noindent \textbf{The Generation stage.} 
Here, the model decodes the output sentence $S$ from $\tilde{E}^{D}$ using a pre-trained BART decoder that is fine-tuned during training with the rest of the model.


\subsection*{Model Training and Inference} 
\noindent \textbf{Training  data preparation.} 
Acquiring training data for our masked conditional sentence generation model  described above 
 is relatively easy as any well-formed sentence can be converted into a training instance. We do this by first identifying a masked word, which can be any verb, adjective, and adverb from the sentence because IEs mostly assume these roles in a sentence. Then, we retrieve the definitions of the masked word from dictionaries. To increase the diversity in definitions and prevent the model from becoming dictionary-specific, we access the masked word's definitions randomly from WordNet \cite{10.1145/219717.219748}, Wiktionary\footnote{https://en.wiktionary.org/}, or Google Dictionary\footnote{https://dictionaryapi.dev/}. Finally, we use a BERT-based \cite{devlin-etal-2019-bert} POS tagger to predict the POS tag for the masked word. 
Inspired by \citet{hegde2020unsupervised}'s way of  improving the fluency of generated sentences  we drop stop words from the input sentences and ask the model to reconstruct them. Hence, in each batch of our  training, 
80\% of the sentences have their stop words removed and 40\% of the sentences have their words lemmatized {(these two operations can happen simultaneously)}. For our case, these sentence corruptions have the additional benefit of allowing the model to  generate more than one word in place of the masked token, which is critical for generating substitutions for several IEs.

\noindent \textbf{Inference.} 
During inference, given an IE, $I$, it is replaced by the masked token $i_{w}$. Then, the POS tag of $i_{w}$ is predicted with a pretrained POS tagger and fed to the model with the masked IE's definition. The model then generates the output $S$ with the masked IE replaced by a literal phrase. It is important to note that the ISP task is performed in a zero-shot manner in that the model is trained to fill in a masked word, but during inference its knowledge and function are transferred to predict the literal meaning of IEs.

\section{The Weakly Supervised Method} \label{sec:supervised_method}
For the low-resource scenario, we use a small parallel dataset 
$\mathbb{P} = \{(I_1, S_1), (I_2, S_2), \cdots , (I_N, S_N)\} = \{\mathbb{I}; \mathbb{S}\}$ of $N$ pairs of sentences, where  ($I_k$, $S_k$) is a pair of idiomatic sentence and its literal  counterpart. 
to create a weakly supervised end-to-end model for ISP. Like \textsc{Bart-Ucd} above, it takes an idiomatic sentence as input (without the IE's definition/identity during training) and generates the entire paraphrased literal sentence as output. Drawing a parallel between ISP and that of machine translation, our weakly supervised approach relies on an iterative back-translation mechanism 
to (generate and) augment the limited training data and improve the performance of a vanilla BART model, which we refer to as \textsc{Bart-Ibt}. 


The limited size of $\mathbb{P}$ prompts us to generate a much larger $\mathbb{I}_M$ by iteratively training two models simultaneously: (1) an ISP model that translates an idiomatic sentence $\mathbb{I}$ to a literal sentence $\hat{\mathbb{S}}$, and (2) an {Idiomatic Sentence Generation (ISG)} model that translates a literal sentence $\mathbb{S}$ into an idiomatic sentence $\hat{\mathbb{I}}$. Note that besides  our main objective of training an ISP model, acquiring a competent ISG model and a larger parallel dataset are both welcome byproducts.

Each training iteration consists of three stages---\textit{Model training},  \textit{Data generation}, and \textit{Data selection}. The iterative process is described in Figure~\ref{fig:supervised_model} and  Algorithm~\ref{algo:supervised}. 

\subsection*{Model Training}

We use the parallel dataset ($\mathbb{P}$ to begin with and the augmented set---described below--- during subsequent iterations) to fine-tune two separate pretrained BART models yielding the ISP and the ISG model. 

\subsection*{Data Generation}
In this stage,  the trained ISG model and the ISP model from the previous stage generate more idiomatic-literal sentence pairs that augment the initial training set. 
First, the ISP model  generates literal counterparts $\hat{\mathbb{S}}_M$ 
for all the idiomatic sentences in $\mathbb{I}_M$. Then the ISG model is used to transform the literal sentences back into the idiomatic form, whose collection is $\hat{\mathbb{I}}_M$.
At the end of this stage, we gather $\hat{\mathbb{S}}_M$ and  $\hat{\mathbb{I}}_M$ to produce the set of candidate pairs $\mathbb{D}_M$ for the next stage. 

\subsection*{Data Selection}
Note that there may be low quality pairs in  $\mathbb{D}_M$ resulting from, e.g., IEs not replaced in the generated literal sentences or IEs  omitted from the  back-translated idiomatic sentences. Toward excluding these pairs  from the collection $\mathbb{D}_M$
 we propose two rules: (1) For any example $(I_M^j, \hat{S}_M^j, \hat{I}_M^j) \in \mathbb{D}_M$, if the literal sentence $\hat{S}_M^j$ still contains the IE in $I_M^j$, the example will be excluded; and (2) for any example $(I_M^j, \hat{S}_M^j, \hat{I}_M^j) \in \mathbb{D}_M$, if the back-transformed idiomatic sentence $\hat{I}_M^j$ is different from the original idiomatic sentence $I_M^j$, the example will be excluded.
After filtering, we get  $\mathbb{D}_M^* \in \mathbb{D}_M$ such that $ = \{\mathbb{I}_M^*; \hat{\mathbb{S}}_M^*\}$, where $\mathbb{I}_M^* \in \mathbb{I}_M$ and $\hat{\mathbb{S}}_M^* \in \hat{\mathbb{S}}_M$. Finally, the parallel dataset $\mathbb{P}$ is enlarged to $\mathbb{P} \cup \mathbb{D}_M^*$. Also,  $\mathbb{I}_M$  is shrunk to $\mathbb{I}_M \setminus \mathbb{I}_M^*$. The enlarged parallel dataset and the updated set of idiomatic sentences are used in the next iteration.

After all the iterations, we obtain  an enlarged parallel dataset with idiomatic/literal sentence pairs and the well-trained models for ISG and ISP.

\begin{algorithm}[t]
\small
    \caption{WeaklySupervisedModel}
    \label{algo:supervised}
    \KwIn{Original parallel dataset $\mathbb{P}$, Idiomatic sentences $\mathbb{I}_M$  and number of iterations $N$} 
    \KwOut{\textbf{ISP} and \textbf{ISG} Model, Enlarged parallel dataset $\mathbb{P}$}
    $\mathbb{P}_1=\mathbb{P}$ , \quad
    $\mathbb{I}_M^1=\mathbb{I}_M$ \;
    \For{$n=1;n \le N$} 
        {
            $\mathbb{D}_M = \varnothing$ \;
            $\textbf{ISP}_n \Leftarrow \textit{TRAIN}(\mathbb{P}_n)$, \quad
            $\textbf{ISG}_n \Leftarrow \textit{TRAIN}(\mathbb{P}_n)$ \;  
            \For{$I_M \in \mathbb{I}_M^n$} 
            { 
                $\hat{S}_M = \textbf{ISP}_n(I_M)$ , \quad
                $\hat{I}_M = \textbf{ISG}_n(\hat{S}_M)$ \;
                $\mathbb{D}_M \Leftarrow \mathbb{D}_M \bigcup \{(I_M, \hat{S}_M, \hat{I}_M)\}$ \;
            }
            $\mathbb{D}_M^* = \varnothing$ \;
            \For{$(I_M, \hat{S}_M, \hat{I}_M) \in \mathbb{D}_M$} 
            { 
                \If{$I_M \neq \hat{S}_M \land \hat{I}_M = I_M$} 
                { 
                $\mathbb{D}_M^* \Leftarrow \mathbb{D}_M^* \bigcup \{(I_M, \hat{S}_M)\}$\; 
                } 
            }
            $\mathbb{I}_M^{n+1} = \mathbb{I}_M^n$ \;
            \For{$(I_M, \hat{S}_M) \in \mathbb{D}_M^*$} 
            { 
                $\mathbb{I}_M^{n+1} \Leftarrow \mathbb{I}_M^{n+1} \setminus I_M$ \;
            } 
            $\mathbb{P}_{n+1} \Leftarrow \mathbb{P}_n \bigcup \mathbb{D}_M^*$ \;
        } 
    return $\textbf{ISP}_{N}, \textbf{ISG}_{N}, \mathbb{P}_{N+1}$\;
\end{algorithm}

\section{Experiments}\label{sec:experiments}
In this section, we evaluate the performances of the proposed \textsc{Bart-Ucd} and \textsc{Bart-Ibt} against  competitive baselines, while later in the paper, we show an application of ISP in a downstream NLP task.

\subsection{Baselines}
We study the following competitive text generation baselines for ISP---the \textbf{Seq2Seq} model \cite{sutskever2014sequence}, the \textbf{Transformer} model \cite{vaswani2017attention}, the copy-enriched Seq2Seq (\textbf{Seq2Seq-copy}) model \cite{jhamtani2017shakespearizing}, the copy-enriched Transformer (\textbf{Transformer-copy}) model \cite{gehrmann2018bottom}, and the \textbf{T5} model \cite{raffel2020exploring}. 
To validate the effectiveness of  \textsc{Bart-Ibt}, we also use a fine-tuned BART (\textbf{BART}) model without back-translation as a baseline.

{Our baselines do not include standard paraphrasing and style-transfer models  due to the lack of a large-scale parallel corpus and the ISP requirement of changing only a single phrase in the sentence. Moreover, we also exclude pre-trained language models   mainly to highlight the overall difficulty of ISP.}

\subsection{Datasets} \label{sec:datasets}
In this section we first introduce the training sets for the proposed  methods followed by the test sets  used by the proposed methods and the baselines.

\subsubsection*{Training Set}

Recall that any corpus of well-formed sentences can be used to train  \textsc{Bart-Ucd}. Accordingly, we choose two large news datasets---AG News \cite{Zhang2015CharacterlevelCN} and  CNN-Dailymail \cite{DBLP:journals/corr/SeeLM17}--- and the GLUE datasets MRPC and COLA \cite{wang-etal-2018-glue}. This choice is guided by the rationale that they are well-formed and less likely to contain IEs  owing to their being sentences from the news and the scientific domain (to minimize the likelihood that the model may  generate IEs). For AG News and CNN-Dailymail, we 
randomly sampled 1 million sentences from each sentence-tokenized dataset. Considering each sentence with a masked word as a data instance, our final training corpus has 1.97 million instances, 11,071 unique masked words, and 17 unique POS tags. Even though including more training instances,  as with all models, can improve the model's performance, we found our current training corpus to yield satisfactory results.  

Toward training \textsc{Bart-Ibt} (i.e., fine-tuning the backbone pretrained BART models for our task), 
we used the parallel dataset  constructed by \citet{zhou2021solving} (henceforth termed  \textbf{PIL}) with a training set of 3,789 manually created idiomatic and literal sentence pairs from a list of 876 IEs and their definitions, with at least 5 idiomatic sentences per IE. 
The  idiomatic sentences (without  literal counterparts) used for \textsc{Bart-Ibt} training are from the MAGPIE corpus \cite{haagsma2020magpie} collected  from the BNC. 
Choosing sentences with IEs used figuratively  yielded 27,582 idiomatic sentences from 1,644 IEs  to form the idiomatic sentence set $\mathbb{I}_M$. Among the 1,644 IEs, 208  overlap with those in PIL. 

All \textbf{baselines} were trained  using only the PIL training set.

\subsubsection*{Test Set.}

For a fair comparison across the methods, we used two types of  test sets to evaluate all the methods. The first was the test split of PIL  for both automatic and manual evaluation. This includes 876 idiomatic-literal sentence pairs with each idiomatic sentence containing a unique IE that occurred in the training set. We leave it to future work to examine generalization to IEs unseen during training.

To afford a different perspective of the models' capabilities with naturally occurring idiomatic instances, we used a second test set  constructed from the MAGPIE dataset (\textbf{MIL}; only for manual evaluation) consisting of 100 idiomatic sentences unseen  in the training set of  \textsc{Bart-Ibt}. The literal counterparts were provided by one annotator and then verified by a second annotator, both native English speakers and proficient users  of IEs and not part of  the research team. To ensure compatibility between  the set of IEs in MIL and PIL, we verified that the same IEs were used in the  idiomatic sentences of the two test sets. 

\begin{table*}[t]
    \small
    \centering
    \begin{tabular}{lcccccccc}
    \thickhline
    \thickhline
        \textbf{Model - ISP} & \textbf{BLEU} & \textbf{ROUGE-1} & \textbf{ROUGE-2} & \textbf{ROUGE-L} & \textbf{METEOR} & \textbf{SARI} & \textbf{GRUEN} & \textbf{PPL} \\
        \thickhline
        
        Seq2Seq & 42.96 & 62.43 & 40.46 & 62.54 & 59.36 & 33.89 & 33.45 & 11.54\\
        
        Transformer & 46.65 & 60.90 & 43.34 & 61.39 & 69.82 & 38.62 & 44.06 & 10.59\\
        
        Seq2Seq-copy & 47.58 & 71.67 & 50.20 & 76.77 & 77.23 & 49.69 & 32.84 & 9.85\\
        
        Transformer-copy & 57.91 & 68.44 & 54.97 & 69.59 & 79.17 & 45.10 & 52.25 & 4.61\\
        
        \hdashline
        T5 & 55.36 & 77.79 & 67.66 & 77.63 & 74.19 & 54.63 & 61.74 & 6.22\\
        
        BART & *78.53\hspace{0.15cm} & 84.64 & 77.21 & 84.95 & 85.36 & 61.82 & *78.03\hspace{0.15cm} & 5.35\\
        
        \textsc{Bart-Ucd} (ours) & 76.58 & *84.92\hspace{0.15cm} & *77.99\hspace{0.15cm} & *85.31\hspace{0.15cm} & *87.80\hspace{0.15cm} & *74.50\hspace{0.15cm} & 77.13 & *5.11\hspace{0.15cm}\\
        
        \textsc{Bart-Ibt} (ours) & \textbf{83.69} & \textbf{87.82} & \textbf{82.47} & \textbf{88.19} & \textbf{87.92} & \textbf{81.39} & \textbf{83.06} & \textbf{3.12}\\
        
        
        
        
        
        
        
        
        \thickhline
        \thickhline
    \end{tabular}
    \caption{Performance comparison for ISP on the PIL test set. The best performance for each metric is in \textbf{bold} and the second best has an asterisk (*).}
    \label{tab:overall2}
\end{table*}

\subsection{Experimental Setup}
\label{sec:setup}

Here we introduce the basic settings for the models.

\noindent \textbf{Unsupervised Method. }
We use the pretrained BART-large model, the BERT-based POS tagger and their respective checkpoints as implemented and hosted by Huggingface's Transformers library. The RoBERTa-based sentence embedding generator and its checkpoint are implemented and hosted by \cite{reimers-2020-multilingual-sentence-bert}.
All of our pretrained transformer-based models are implemented and hosted by Huggingface's Transformers library. Specifically, for \textsc{Bart-Ucd}, we use a pretrained BART-large\footnote{https://huggingface.co/facebook/bart-large-cnn} model and BERT-based POS tagger\footnote{https://huggingface.co/vblagoje/bert-english-uncased-finetuned-pos}.
The RoBERTa-based sentence embedding generator \cite{reimers-2020-multilingual-sentence-bert} is implemented the Sentence-Transformers library\footnote{https://huggingface.co/sentence-transformers/roberta-base-nli-stsb-mean-tokens}. The model was trained with an initial learning rate of $1e-5$, a linear scheduler with 20,000 warm-up steps, and a batch size of $16$ for $3$ epochs. The maximum sequence length was set to be $128$. During inference, we used a beam search with $5$ beams with top-\textit{k}  set to $100$ and top-\textit{p} set to $0.5$. The other hyper-parameters were set to their default values. The model is trained and tested on a single machine with an Intel\textsuperscript{\tiny\textregistered} Core\textsuperscript{\tiny\texttrademark} i9-9900K processor and a single NVIDIA\textsuperscript{\tiny\textregistered} GeForce\textsuperscript{\tiny\textregistered} RTX 2080 Ti graphic card. There are 423M parameters in  \textsc{Bart-Ucd}.

\noindent \textbf{Weakly Supervised Method. }
We used two independent pretrained BART-large models as the ISP model and the ISG model in \textsc{Bart-Ibt}. These pretrained  models were also implemented as hosted by Huggingface's Transformers library. The maximum length for a sentence, the learning rate and the number of iterations were 128,  $5e-5$, and 5 respectively. The other hyper-parameters were  their  default values.  This model is trained and inferenced on Google Colab platform. Each Bart-large model has 406M parameters.

\noindent \textbf{Baselines. }
For the Seq2Seq, the Transformer, the Seq2Seq-copy, and the Transformer-copy, we followed the experimental settings described in \cite{zhou2021solving, zhou-etal-2021-pie}; the baseline pretrained BART model is identical to that used in \textsc{Bart-Ibt}, and the T5 model is  that hosted by Huggingface and trained under the same settings as the BART model. For the baseline pretrained BART model (also from Huggingface's Transformers), the maximum length for a sentence and the learning rate were set to 128 and  $5e-5$, respectively. 
The model was trained for 5 epochs. During inference, we used a beam search with 5 beams with top-\textit{k} set to 100 and top-\textit{p} set to 0.5. The other hyper-parameters were set to their default values. This model is trained and inferenced on Google Colab platform. T5 model is trained in the same way.

\begin{table*}[t]
    \small
    \centering
    \begin{tabular}{c l l}
    \thickhline
    \multicolumn{2}{c}{\textbf{Literal sentence}} & But dear Caroline's got an almighty hangover, \textbf{\color{blue}{very ill}}, so I brought him over on the back of the bike. \\
    \hline
    \multicolumn{2}{c}{\textbf{Idiomatic sentence}} & But dear Caroline's got an almighty hangover, \textbf{\color{red}{sick as a dog}}, so I brought him over on the back of the bike. \\
    \hline
    \hline
    \multicolumn{1}{c|}{\multirow{8}{*}{\textbf{ISP}}} &
    \textbf{Seq2Seq} & but caroline got, as as, so I brought him over . \\
    \multicolumn{1}{c|}{} & \textbf{Transformer} & but dear caroline's got an almighty hangover, \textbf{\color{red}{sick as a dog}}, so I brought him over. \\
     \multicolumn{1}{c|}{} & \textbf{Seq2Seq-copy} & but dear caroline's got an an, sick as as, so I brought him over on on the back.\\
     \multicolumn{1}{c|}{} & \textbf{Transformer-copy} & but dear caroline's got an almighty hangover, \textbf{\color{red}{sick as a dog}}, so I brought him over on the back of the bike. \\
     \multicolumn{1}{c|}{} & \textbf{T5} & But dear Caroline's got an almighty hangover, \textbf{\color{red}{sick as a dog}}, so I brought him over on the back of the bike. \\
     \multicolumn{1}{c|}{} & \textbf{BART} & But dear Caroline's got an almighty hangover, \textbf{\color{red}{sick as a dog}}, so I brought him over on the back of the bike. \\
     \multicolumn{1}{c|}{} & \textbf{\textsc{Bart-Ibt}} (Ours) & But dear Caroline's got an almighty hangover, \textbf{\color{blue}{feeling sick}}, so I brought him over on the back of the bike. \\
     \multicolumn{1}{c|}{} & \textbf{\textsc{Bart-Ucd}} (Ours) & But dear Caroline's got an almighty hangover, \underline{\color{olive}{sick}}, so I brought him over on the back of the bike. \\
    \thickhline
    \end{tabular}
    \caption{A sample of generated literal sentences. Text in \textit{\textbf{\color{red}{bold and italics red}}}  represents the IEs, text in \textbf{\color{blue}{bold blue}}  represents the correct literal counterparts in the outputs, and text in \underline{\color{olive}{underlined olive}} represents the poorly generated literal phrases.}
    \label{tab:sample1}
\end{table*}

\begin{table*}
    \small
    \centering
    \begin{tabular}{lcccccccc|cccccccc}
    \thickhline
         & \multicolumn{8}{c|}{\textbf{PIL Test Set}} & \multicolumn{8}{c}{\textbf{MIL Test Set}} \\ \cmidrule{2-17} \small \textbf{Model} & \multicolumn{2}{c}{\textbf{Meaning}} & \multicolumn{2}{c}{\textbf{Target}} & \multicolumn{2}{c}{\textbf{Fluency}} & \multicolumn{2}{c|}{\textbf{Overall}} & \multicolumn{2}{c}{\textbf{Meaning}} & \multicolumn{2}{c}{\textbf{Target}} & \multicolumn{2}{c}{\textbf{Fluency}} & \multicolumn{2}{c}{\textbf{Overall}} \\ \cmidrule(lr){2-3} \cmidrule(lr){4-5} \cmidrule(lr){6-7} \cmidrule(lr){8-9} \cmidrule(lr){10-11} \cmidrule(lr){12-13} \cmidrule(lr){14-15} \cmidrule(lr){16-17} & \multicolumn{1}{c}{\textbf{Scr.}} & \multicolumn{1}{c}{\textbf{Agr.}} & \multicolumn{1}{c}{\textbf{Scr.}} & \multicolumn{1}{c}{\textbf{Agr.}} & \multicolumn{1}{c}{\textbf{Scr.}} & \multicolumn{1}{c}{\textbf{Agr.}} & \multicolumn{1}{c}{\textbf{Scr.}} & \multicolumn{1}{c|}{\textbf{Agr.}} & \multicolumn{1}{c}{\textbf{Scr.}} & \multicolumn{1}{c}{\textbf{Agr.}} & \multicolumn{1}{c}{\textbf{Scr.}} & \multicolumn{1}{c}{\textbf{Agr.}} & \multicolumn{1}{c}{\textbf{Scr.}} & \multicolumn{1}{c}{\textbf{Agr.}} & \multicolumn{1}{c}{\textbf{Scr.}} & \multicolumn{1}{c}{\textbf{Agr.}} \\
        \thickhline
        \footnotesize BART & 0.73 & 0.88 & 2.56 & 0.57 & \textbf{3.85} & 0.80 & 1.30 & 0.56 & 0.53 & 0.92 & 1.70 & 0.54 & 2.37 & 0.80 & 0.92 & 0.58\\
        
        \footnotesize \textsc{Bart-Ucd} & 0.48 & 0.74 & 2.25 & 0.42 & 3.43 & 0.59 & 1.13 & 0.56 & 0.64 & 0.74 & 2.21 & 0.42 & 3.16 & 0.57 & 0.98 & 0.54\\
        
        \footnotesize \textsc{Bart-Ibt} & \textbf{0.81} & 0.83 & \textbf{3.11} & 0.47 & \textbf{3.85} & 0.80 & \textbf{1.63} & 0.47 & \textbf{0.80} & 0.89 & \textbf{2.48} & 0.47 & \textbf{3.36} & 0.63 & \textbf{1.28} & 0.47 \\
        \thickhline
    \end{tabular}
    \caption{Human evaluation results for ISP based on the PIL and MIL test sets. The best performance is in bold. \textbf{Scr.} represents the humam evaluation scores and \textbf{Agr.} represents the human evaluation inter-annotator agreement.}
    \label{tab:manual}
\end{table*}

\begin{table*}[!t]
\small
    \centering
    \begin{tabular}{l m{12cm}}
    \thickhline
    \textbf{English Idiomatic Sentence} & I do not know if she is present , but I would like to {\color{red}{\textbf{pass on}}} my deepest condolences to her . \\
    \textbf{German Translation} & Ich wei\ss{} nicht, ob sie anwesend ist, aber ich m\"{o}chte mein tiefstes Beileid \\
    \hline
    \textbf{English Literal Sentence} & I do not know if she is present , but I would like to {\color{blue}{\textbf{express}}} my deepest condolences to her . \\
    \textbf{German Translation} & Ich wei\ss{} nicht, ob sie anwesend ist, aber ich m\"{o}chte ihr mein tiefstes Beileid aussprechen \\
    \thickhline
    \end{tabular}
    \caption{Example that shows how ISP helps En-De machine translation.}
    \label{tab:mt_example}
\end{table*}

\subsection{Evaluation Metrics}

\noindent \textbf{Automatic Evaluation. } We used metrics widely used in  text generation tasks such as paraphrasing and style transfer---ROUGE \cite{lin2004rouge}, BLEU \cite{papineni2002bleu} and METEOR \cite{lavie2007meteor}---to compare  the generated sentences with the references. Due to the similarity between ISP and text simplification, we also used SARI \cite{xu2016optimizing}, the metric for  text simplification. To measure linguistic quality, we use a pretrained GPT-2 \cite{radford2019language} to calculate perplexity scores and a recently proposed measure of linguistic quality, GRUEN \cite{zhu2020gruen}. These scores were collected on the PIL test set.

\noindent \textbf{Human Evaluation.} For a qualitative measure of ISP 
we use human evaluation to complement the automatic evaluation.
We used  100 instances from the PIL test set and the entire MIL test set, and collected the outputs from the 3 best methods ranked by automatic evaluation. For each output sentence, two native English speakers, who were blind to the systems being compared,  were asked to rate the output sentences with respect to  meaning, style and fluency  using the following scoring criteria: \\
\noindent (1) \textbf{Meaning preservation} measures on a binary scale how well the meaning of the input is preserved in the output.\\
\noindent (2) \textbf{Target inclusion} indicates on a scale of 1-4  whether the correct literal phrase was used in the output (1: the target phrase was not included in the output at all, 2: partial inclusion, 3:  complete inclusion of a different phrase but with similar meaning with the target, and  4: complete inclusion).
\\
\noindent (3) \textbf{Fluency} evaluates the naturalness and the readability of the output, including the appropriate use of the verb tense, noun and pronoun forms, on a scale of 1 to 4, ranging from ``highly nonfluent'' to ``very fluent.''
\\
\noindent (4) \textbf{Overall} evaluates the overall quality of output on a scale of 0 to 2 like that used to evaluate paraphrases  \cite{iyyer-etal-2018-adversarial}, jointly capturing meaning preservation and fluency: a score of 0 for a sentence that was clearly wrong, grammatically incorrect or does not preserve meaning; a score of 1 for a sentence with minor grammatical errors or   meaning  largely preserved from the original but not completely; score 2 denotes that the sentence is  grammatically correct and the meaning is preserved.
\\

\section{Results and Discussion}


\noindent \textbf{\textsc{Bart-Ucd}. } As shown in Table~\ref{tab:overall2}, without training on PIL, \textsc{Bart-Ucd} outperforms the  supervised baselines in 6 out of 8 metrics for the task of ISP and achieves a competitive performance with the strongly supervised BART outperforming it   by 2.44 (METEOR) and 12.68 (SARI) points.

\noindent \textbf{\textsc{Bart-Ibt}. } As shown in Table~\ref{tab:overall2}, 
\textsc{Bart-Ibt} achieves the best performance across all metrics, even though its actual performance may be underrepresented by the automatic metrics that fail to capture meaning equivalences despite differences in surface form. 

\noindent \textbf{Model Comparison. } Overall, the pretrained BART model, our \textsc{Bart-Ibt} and \textsc{Bart-Ucd} perform competitively on ISP going by the metrics METEOR and ROUGE-1. However, 
a qualitative analysis shows that BART tends to copy the input sentence in the output  15\% of time and on an average only modifies 9\% of the tokens from the input sentences, suggesting an \textit{overrepresentation} of its performance by the automatic metrics. On the contrary, while being good at copying context words (a desirable feature),  \textsc{Bart-Ibt} outperforms the other models   
showing the best SARI score (a measure of the novelty in the generated output compared to the input). This underscores the importance of  the iterative back-translation mechanism without which the  performance gains would have been impossible.
{Moreover, we note that BART performs better on PIL while \textsc{Bart-Ucd} performs better on MIL. A plausible explanation for this divergence is that  
PIL is synthetically created idiomatic sentences whereas  MIL is in-the-wild ones. Thus, MIL is an out of distribution, yet more general test data for BART that was trained on PIL. However, \textsc{Bart-Ucd}, being agnostic to PIL, 
is indifferent to the distribution shift in MIL. 
}

\noindent \textbf{Human Evaluation. } 
The results of human evaluation are presented in Table~\ref{tab:manual}. We note  that the output of \textsc{Bart-Ibt} was rated the best across all the dimensions. 
It appears that vanilla BART  performs on par with \textsc{Bart-Ibt} in meaning preservation and fluency. 
However, BART's tendency to copy its input artificially inflates its meaning preservation and fluency scores. 
It is worth noting that when tested on MIL, both  \textsc{Bart-Ucd} and \textsc{Bart-Ibt} outperform the pretrained BART in corresponding tasks, which speaks of the generalizability of \textsc{Bart-Ucd} and \textsc{Bart-Ibt} to the naturally occurring idiomatic sentences in MIL.  Averaged over the four dimensions, the inter-annotator agreement score  was 0.58 for \textsc{Bart-Ucd} and 0.62 for \textsc{Bart-Ibt}.

\noindent \textbf{Error Analysis.} 
The main challenge for all the models seems to be generating long informative literal phrases based on the correct sense of the IE.  For example, \textsc{Bart-Ibt} replaces the IE \textit{blow hot and cold} with \textit{fluctuate}, which is inaccurate and the reason for annotators to diverge on Target inclusion and Overall scores. 

\noindent \textbf{Byproducts from \textsc{Bart-Ibt}.} The back-translation mechanism used in \textsc{Bart-Ibt} leads to an ISG model (in addition to the ISP model) after training. To evaluate its competence, we perform the same automatic and human evaluations against the same set of baseline models. From the results, we found that  \textsc{Bart-Ibt} outperforms  all the baselines across all automatic metrics by wide margins, ranging from 11.76 higher in BLEU, 12.92 higher in ROUGE-2 and 16.32 higher in  SARI  over  the  next  best  model, while achieving the best performance across all human metrics as well. Besides, we also obtain a large scale parallel dataset to spur future explorations, which includes 1,169 IEs with 15,627 idiomatic/literal sentence pairs. 

Table~\ref{tab:sample1} shows the  sentences generated by the different models for the same IE input. More examples are available in the Appendix.

\section{Application}
The challenges posed by IEs to machine translation owing to inadequate handling of  non-compositional phrases  has been documented  by \citet{fadaee2018examining} who also provide a challenge set of idiomatic sentences. Here we explore the extent to which using ISP as a preprocessing step to remove all the IEs from the input sentences  can reduce the negative influence of IEs  in machine translation. Performing ISP as a preprocessing step is  inexpensive  and flexible since it does not require the expansive development or retraining of new models to handle IE specifically and can be widely used in any downstream application.

Specifically, we use \textsc{Bart-Ibt}  to first transfer the idiomatic sentences into literal sentences in the source language. Then, we use a state-of-the-art NMT system to translate the resulting literal sentences into the target language. 
We run experiments using the challenge test set for English-to-German translation  constructed by \cite{fadaee2018examining} that consists of idiomatic sentences in English and their corresponding translations in German. There were 1,500 En-De pairs in the test set, using a total of 132 IEs. We used a pre-trained  mBART \cite{liu2020multilingual} as the NMT system with all the parameters  set to their default values. 

As a result of  the pre-processing using \textsc{Bart-Ibt}, the BLEU score on the challenge set improved from $10.1$ to $10.7$, which shows the effectiveness of the ISP in a downstream NLP application. Though this improvement may not seem substantial, we stress that this gain comes with just a preprocessing step and no other change in training.   Table~\ref{tab:mt_example} shows an example of how ISP helps the translation of idiomatic sentences. In the original translation, the main verb \textit{aussprechen} is missing. However, when the IE `pass on' is replaced with `expressed', the translation is complete.

\section{Conclusion}
\label{sec:conclusion}

 In this paper, we studied the task of idiomatic sentence paraphrasing (ISP) in a zero- and low-resource setting. We proposed an unsupervised method that utilizes contextualized word embeddings and word definition sentence embeddings for ISP. In addition, we explored the use of a weakly supervised method based on an iterative back-translation mechanism. Our experiments and analyses demonstrate that unsupervised  and weakly supervised methods show  competitive paraphrasing  performance in low-resource settings, with the weakly supervised method  outperforming available baseline methods in all evaluation dimensions. Furthermore, the weakly supervised approach yields an ISG model and a large-scale parallel dataset.
 
 The limitations of this study include conducting the study without a large   parallel dataset of high quality, assuming one sense for IEs \cite{hummer2006polysemy}, limiting each sentence to have only one IE and using a list of IEs that did not account for the diversity of World Englishes \cite{pitzl2016world}. Future work should address these limitations. 

\section{Acknowledgments}
The research reported here was supported by the Institute of Education Sciences, U.S. Department of Education through  Grant number R305A180211 to the Board of Trustees of the University of Illinois. The opinions expressed are those of the authors and do not represent views of the Institute or the U.S. Department of Education.

\bibliography{aaai22}
\clearpage

\appendix

\section{Appendix}
\label{sec:appendix}

\subsection{Evaluation Results on ISG model from \textsc{Bart-Ibt}}
Although the main objective for \textsc{Bart-Ibt} is to train a better ISP model with the limited parallel data resources, we also obtain a quite competent ISG model as a welcome byproduct of the back-translation process. For comparison, we trained the same set of baseline models to perform the ISG task. As shown in Table \ref{tab:ISG_res},  our \textsc{Bart-Ibt} outperforms all the baselines across all metrics by wide margins ranging from 11.76 higher in BLEU, 12.92 higher in ROUGE-2 and 16.32 higher in SARI over the next best model, achieving near-perfect performances in these metrics.

To further ensure the quality of the generated sentence from the ISG model, we also conducted human evaluation using the same scoring metrics as the ISP model. As shown in Table \ref{tab:manual_ISG_res}, \textsc{Bart-Ibt}'s ISG model outperforms the Pipeline and Vanilla BART in all human evaluation metrics. To put the human evaluation performance into context, we also supply the inter-annotator agreement for all the criteria for ISG and ISP as shown in Table \ref{tab:agreement}.

Despite the great performance, after observing error cases, we found that sometimes \textsc{Bart-Ibt}'s ISG model appears to rely on the occurrence of some trigger words for choosing some idioms, instead of relying on  on the context and the meaning of the whole sentence, e.g. \textit{eat} is frequently replaced with \textit{chow down}. Besides, when using IEs in the output sentence, \textsc{Bart-Ibt} attempts to use an IE with a similar meaning instead of the target IE.

\begin{table*}[t]

    \centering
    \begin{tabular}{lcccccccc}
    \thickhline
    \thickhline
        \textbf{Model - ISG} & \textbf{BLEU} & \textbf{ROUGE-1} & \textbf{ROUGE-2} & \textbf{ROUGE-L} & \textbf{METEOR} & \textbf{SARI} & \textbf{GRUEN} & \textbf{PPL} \\
        \thickhline
        
        Seq2Seq & 25.16 & 48.26 & 22.90 & 47.21 & 41.46 & 24.13 & 32.25 & 9.24\\
        
        Transformer & 45.58 & 60.22 & 42.82 & 60.59 & 68.68 & 36.67 & 44.05 & 6.00\\
        
        Seq2Seq-copy & 38.02 & 66.11 & 40.37 & 74.04 & 68.21 & 43.02 & 27.79 & 5.43\\
        
        Transformer-copy & 59.56 & 68.34 & 55.72 & 69.38 & 79.53 & 39.93 & 59.27 & 4.12\\
        
        T5 & 55.66 & 77.49 & 67.79 & 77.24 & 74.75 & 59.78 & 53.46 & 5.03\\
        
        Pipeline & 65.56 & 74.44 & 62.96 & 74.56 & 78.02 & 67.64 & 67.27 & 3.4 \\
        \hdashline
        BART & 79.32 & 83.95 & 77.16 & 84.20 & 83.41 & 62.30 & 77.49 & 3.88 \\

        \textsc{Bart-Ibt} (ours) & \textbf{91.08} & \textbf{93.01} & \textbf{90.08} & \textbf{93.19} & \textbf{92.86} & \textbf{83.87} & \textbf{89.13} & \textbf{3.01}\\
        
        \thickhline
        \thickhline
    \end{tabular}
    \caption{Performance comparison for ISG on the original test set (PIL). The best performance for each metric in in \textbf{bold}.}
    \label{tab:ISG_res}
    
    \centering
    \begin{tabular}{lcccc|cccc}
    \thickhline
         & \multicolumn{4}{c}{\textbf{PIL Test Set}} & \multicolumn{4}{c}{\textbf{MIL Test Set}} \\ 
         \cmidrule{2-5}\cmidrule{6-9} \textbf{Model} & \textbf{Meaning} & \textbf{Target}& \textbf{Fluency}& \textbf{Overall} & \textbf{Meaning} & \textbf{Target}& \textbf{Fluency} & \textbf{Overall} \\ 
        \thickhline
        \footnotesize Pipeline & 0.59  & 2.12  & 3.45 & 1.12 & 0.50 & 1.67 & 2.57  & 0.88 \\
        \footnotesize BART & 0.76  & 2.52  & 3.83 & 1.32  &0.46 & 1.52 &   2.07 &  0.83 \\
        \footnotesize \textsc{Bart-Ibt} & \textbf{0.99}  & \textbf{3.99}  & \textbf{3.97}  & \textbf{1.94}  & \textbf{0.88} & \textbf{3.35}  & \textbf{3.51}  & \textbf{1.74} \\
        \thickhline
    \end{tabular}
    \caption{Human evaluation results based on PIL test set and MIL test set for ISG.}
    \label{tab:manual_ISG_res}
    
    \small
    \centering
    \begin{tabular}{lcccccccc|cccccccc}
    \thickhline
         & \multicolumn{8}{c|}{\textbf{PIL Test Set}} & \multicolumn{8}{c}{\textbf{MIL Test Set}} \\ \cmidrule{2-17} \small \textbf{Model} & \multicolumn{2}{c}{\textbf{Meaning}} & \multicolumn{2}{c}{\textbf{Target}} & \multicolumn{2}{c}{\textbf{Fluency}} & \multicolumn{2}{c|}{\textbf{Overall}} & \multicolumn{2}{c}{\textbf{Meaning}} & \multicolumn{2}{c}{\textbf{Target}} & \multicolumn{2}{c}{\textbf{Fluency}} & \multicolumn{2}{c}{\textbf{Overall}} \\ \cmidrule(lr){2-3} \cmidrule(lr){4-5} \cmidrule(lr){6-7} \cmidrule(lr){8-9} \cmidrule(lr){10-11} \cmidrule(lr){12-13} \cmidrule(lr){14-15} \cmidrule(lr){16-17} & \multicolumn{1}{c}{\textbf{ISG}} & \multicolumn{1}{c}{\textbf{ISP}} & \multicolumn{1}{c}{\textbf{ISG}} & \multicolumn{1}{c}{\textbf{ISP}} & \multicolumn{1}{c}{\textbf{ISG}} & \multicolumn{1}{c}{\textbf{ISP}} & \multicolumn{1}{c}{\textbf{ISG}} & \multicolumn{1}{c|}{\textbf{ISP}} & \multicolumn{1}{c}{\textbf{ISG}} & \multicolumn{1}{c}{\textbf{ISP}} & \multicolumn{1}{c}{\textbf{ISG}} & \multicolumn{1}{c}{\textbf{ISP}} & \multicolumn{1}{c}{\textbf{ISG}} & \multicolumn{1}{c}{\textbf{ISP}} & \multicolumn{1}{c}{\textbf{ISG}} & \multicolumn{1}{c}{\textbf{ISP}} \\
        \thickhline
        \footnotesize Pipeline & 0.83 & - & 0.64 & - & 0.68 & - & 0.72 & - & 0.88 & - & 0.67 & - & 0.70 & - & 0.75 & -\\
        
        \footnotesize BART & 0.93 & 0.88 & 0.76 & 0.57 & 0.85 & 0.80 & 0.75 & 0.56 & 0.94 & 0.92 & 0.77 & 0.54 & 0.89 & 0.80 & 0.77 & 0.58\\
        
        \footnotesize \textsc{Bart-Ucd} & - & 0.74 & - & 0.42 & - & 0.59 & - & 0.56 & - & 0.74 & - & 0.42 & - & 0.57 & - & 0.54\\
        
        \footnotesize \textsc{Bart-Ibt} & 0.95 & 0.83 & 0.98 & 0.47 & 0.94 & 0.80 & 0.92 & 0.47 & 0.95 & 0.89 & 0.82 & 0.47 & 0.86 & 0.63 & 0.84 & 0.47 \\
        \thickhline
    \end{tabular}
    \caption{Human evaluation inter-annotator agreement on all the criteria based on PIL test set and MIL test set for the two ISG and ISP.}
    \label{tab:agreement}
\end{table*}

\subsection{More Examples}
\label{sec:more_examples}

In Table~\ref{tab:outputs1} and Table~\ref{tab:outputs2}, we provide more generated examples for the two tasks giving a comparative view of the different models under certain attribute constraints (e.g., high/low compositionality).

\subsection{Task Comparison}

From Table \ref{tab:overall2}, the performance on ISG is better than the performance on ISP. This is due to the limitation of automatic metrics and the nature of the usage of IEs and their literal counterparts. Compared with the usage of IEs, the use of their literal counterparts can be much more flexible. Therefore, a good literal counterpart for a given idiomatic sentence is not necessarily same with the reference, which means that a higher automatic evaluation score may not be a sign of better paraphrasing result because some synonymous idioms or literal phrases will be ignored while they are still appropriate.

\begin{table*}[!b]
\small
    \centering
    \begin{tabular}{l m{12cm}}
    \thickhline
    \textbf{Attribute} &\textbf{multiple IEs} \\
    \hline
    \textbf{Literal sentence} & Without the main character , the show would have been \textbf{\color{blue}{in a state of confusion}}. \\
    \hline
    \textbf{Reference} & Without the main character , the show would have been \textbf{\color{red}{at sixes and sevens}}. \\
    \hline
    \textbf{Seq2Seq with copy} & without the main character , the show would have at at and and . . \\
    \textbf{Transformer with copy} & without the main character , the show would have been \textbf{\color{green}{at arm 's door}}. \\
    \textbf{Pretrained BART} & Without the main character, the show would have been \textbf{\color{green}{at a dead end}}. \\
    \textbf{Pipeline} & Without the main character, the show would have been \textbf{\color{green}{shades of grey}}. \\
    \textbf{\textsc{Bart-Ibt}} & Without the main character, the show would have been \textbf{\color{red}{at sixes and sevens}}. \\
    \hline
    \hline
    \textbf{Attribute} & \textbf{high non-compositionality} \\
    \hline
    \textbf{Literal sentence} & You can't \textbf{\color{blue}{delay making a decision}} any longer , you need to make up your mind . \\
    \hline
    \textbf{Reference} & You can't \textbf{\color{red}{sit on the fence}} any longer , you need to make up your mind . \\
    \hline
    \textbf{Seq2Seq with copy} & you can't sit \textbf{\color{green}{sit the fence}} any , , you need to to up your .
 \\
    \textbf{Transformer with copy} & you can't \textbf{\color{blue}{delay making a decision}} any longer , you need to make up your mind . \\
    \textbf{Pretrained BART} & You can't \textbf{\color{blue}{delay making a decision}} any longer, you need to make up your own mind. \\
    \textbf{Pipeline} & You can't \textbf{\color{blue}{delay making a decisione}} any longer, you make your mind. \\
    \textbf{\textsc{Bart-Ibt}} & You can't \textbf{\color{red}{sit on the fence}} any longer , you need to make up your mind. \\
    \hline
    \hline
    \textbf{Attribute} & \textbf{low non-compositionality} \\
    \hline
    \textbf{Literal sentence} & But dear Caroline 's got an almighty hangover , \textbf{\color{blue}{very ill}} , so I brought him over on the back of the bike to see it again . \\
    \hline
    \textbf{Reference} & But dear Caroline 's got an almighty hangover , \textbf{\color{red}{sick as a dog}} , so I brought him over on the back of the bike to see it again . \\
    \hline
    \textbf{Seq2Seq with copy} & But dear Caroline's got an, \textbf{\color{blue}{very ill}}, so I brought him over over to see \\
    \textbf{Transformer with copy} & but dear caroline's got an almighty hangover, \textbf{\color{blue}{very ill}}, so I brought him over on the back of the bike to see it again. \\
    \textbf{Pretrained BART} & But dear Caroline's got an almighty hangover, \textbf{\color{blue}{very ill}}, so I brought him over on the back of the bike to see it again. \\
    \textbf{Pipeline} & But dear Caroline's got an almighty hangover, very \textbf{\color{green}{under the weather}}, so I brought him over on the back of the bike to see it again. \\
    \textbf{\textsc{Bart-Ibt}} & But dear Caroline's got an almighty hangover, \textbf{\color{red}{sick as a dog}}, so I brought him over on the back of the bike to see it again. \\
    \hline
    \hline
    \textbf{Attribute} & \textbf{high non-compositionality} \\
    \hline
    \textbf{Literal sentence} & In his absence she 'd been as nervy as a wildcat , jumping a mile every time someone spoke to her or touched her on the shoulder , expecting him to turn up \textbf{\color{blue}{unexpectedly}} as he 'd made a habit of doing . \\
    \hline
    \textbf{Reference} & In his absence she 'd been as nervy as a wildcat , jumping a mile every time someone spoke to her or touched her on the shoulder , expecting him to turn up \textbf{\color{red}{out of the blue}} as he 'd made a habit of doing . \\
    \hline
    \textbf{Seq2Seq with copy} & In his absence she'd been as nervy as a a, jumping a mile someone spoke to her, expecting him to turn up up as habit of doing . \\
    \textbf{Transformer with copy} & In his absence she 'd been as nervy as a wildcat , jumping a mile every time someone spoke to her or touched her on the shoulder , expecting him to turn up \textbf{\color{blue}{unexpectedly}} as he 'd made a habit of doing . \\
    \textbf{Pretrained BART} & In his absence she 'd been as nervy as a wildcat , jumping a mile every time someone spoke to her or touched her on the shoulder , expecting him to turn up \textbf{\color{green}{on the doorstep}} as he 'd made a habit of doing . \\
    \textbf{Pipeline} & in his absence she 'd been as nervy as a wildcat, jumping a mile every time someone spoke to her or touched her on the shoulder, expecting him to turn up \textbf{\color{blue}{unexpectedly}} as he 'd made a habit of doing. \\
    \textbf{\textsc{Bart-Ibt}} & In his absence she 'd been as nervy as a wildcat, jumping a mile every time someone spoke to her or touched her on the shoulder, expecting him to turn up \textbf{\color{red}{out of the blue}} as he 'd made a habit of doing. \\
    \thickhline
    \end{tabular}
    \caption{Samples of generated idiomatic sentences. Text in \textbf{\color{red}{bold red}}  represents the idiomatic expressions correctly included in the outputs; text in \textbf{\color{blue}{bold blue}}  represents the literal counterparts in the input sentences. text in \textbf{\color{green}{green}} represents the idioms that are poorly generated.}
    \label{tab:outputs1}
\end{table*}

\begin{table*}[!b]
\small
    \centering
    \begin{tabular}{l m{12cm}}
    \thickhline
    \textbf{Attribute} &\textbf{multiple meaning} \\
    \hline
    \textbf{Idiomatic sentence} & Without the main character , the show would have been \textbf{\color{red}{at sixes and sevens}}. \\
    \hline
    \textbf{Reference} & Without the main character , the show would have been \textbf{\color{blue}{in a state of confusion}}. \\
    \hline
    \textbf{Seq2Seq with copy} & without the main character , the show would have been at a a of confusion . \\
    \textbf{Transformer with copy} & without the main character , the show would have been in complete disaster . \\
    \textbf{Pretrained BART} & Without the main character, the show would have been \textbf{\color{green}{inconsistent}}. \\
    \textbf{\textsc{Bart-Ucd}} & Without the main character, the show would have been \textbf{\color{green}{muddled}}. \\
    \textbf{\textsc{Bart-Ibt}} & Without the main character , the show would have been \textbf{\color{blue}{at a state of confusion}}. \\
    \hline
    \hline
    \textbf{Attribute} & \textbf{high non-compositionality} \\
    \hline
    \textbf{Idiomatic sentence} & You can't \textbf{\color{red}{sit on the fence}} any longer , you need to make up your mind . \\
    \hline
    \textbf{Reference} & You can't \textbf{\color{blue}{delay making a decision}} any longer , you need to make up your mind . \\
    \hline
    \textbf{Seq2Seq with copy} & you can't \textbf{\color{green}{delay making}} any any any , you need to make your your mind . \\
    \textbf{Transformer with copy} & you ca n't sit on the troublesome any longer , you need to make your mind . \\
    \textbf{Pretrained BART} & You can't \textbf{\color{blue}{be indecisive}} any longer, you need to make up your mind. \\
    \textbf{\textsc{Bart-Ucd}} & You can't \textbf{\color{blue}{be indecisive}} any longer, you need to make up your mind. \\
    \textbf{\textsc{Bart-Ibt}} & You can't \textbf{\color{blue}{sit and watch}} any longer, you need to make up your mind. \\
    \hline
    \hline
    \textbf{Attribute} & \textbf{low non-compositionality} \\
    \hline
    \textbf{Idiomatic sentence} & But dear Caroline 's got an almighty hangover , \textbf{\color{red}{sick as a dog}} , so I brought him over on the back of the bike to see it again . \\
    \hline
    \textbf{Reference} & But dear Caroline 's got an almighty hangover , \textbf{\color{blue}{very ill}} , so I brought him over on the back of the bike to see it again . \\
    \hline
    \textbf{Seq2Seq with copy} & but dear caroline's got an an, sick as as, so I brought him over on on the back. \\
    \textbf{Transformer with copy} & but dear caroline's got an almighty hangover, \textbf{\color{red}{sick as a dog}}, so I brought him over on the back of the bike to see it again. \\
    \textbf{Pretrained BART} & But dear Caroline's got an almighty hangover, \textbf{\color{red}{sick as a dog}}, so I brought him over on the back of the bike to see it again. \\
    \textbf{\textsc{Bart-Ucd}} & But dear Caroline's got an almighty hangover, \textbf{\color{green}{sick}}, so I brought him over on the back of the bike to see it again. \\
    \textbf{\textsc{Bart-Ibt}} & But dear Caroline's got an almighty hangover, \textbf{\color{blue}{feeling sick}}, so I brought him over on the back of the bike to see it again. \\
    \hline
    \hline
    \textbf{Attribute} & \textbf{high non-compositionality} \\
    \hline
     \textbf{Idiomatic sentence} & In his absence she 'd been as nervy as a wildcat , jumping a mile every time someone spoke to her or touched her on the shoulder , expecting him to turn up \textbf{\color{red}{out of the blue}} as he 'd made a habit of doing . \\
    \hline
    \textbf{Reference} & In his absence she 'd been as nervy as a wildcat , jumping a mile every time someone spoke to her or touched her on the shoulder , expecting him to turn up \textbf{\color{blue}{unexpectedly}} as he 'd made a habit of doing . \\
    \hline
    \textbf{Seq2Seq with copy} & In his absence she'd been as nervy as a a, jumping a mile someone spoke to her, expecting him to turn up up out of the the . \\
    \textbf{Transformer with copy} & in his absence she 'd been as nervy as a wildcat, jumping a mile every time someone spoke to her or touched her on the shoulder, expecting him to turn up \textbf{\color{red}{out of the blue}} as he 'd made a habit of doing. \\
    \textbf{Pretrained BART} & In his absence she 'd been as nervy as a wildcat , jumping a mile every time someone spoke to her or touched her on the shoulder , expecting him to turn up \textbf{\color{red}{out of the blue}} as he 'd made a habit of doing . \\
    \textbf{\textsc{Bart-Ucd}} & In his absence she 'd been as nervy as a wildcat, jumping a mile every time someone spoke to her or touched her on the shoulder, expecting him to turn up \textbf{\color{blue}{unexpectedly}} as he 'd made a habit of doing.\\
    \textbf{\textsc{Bart-Ibt}} & In his absence she 'd been as nervy as a wildcat, jumping a mile every time someone spoke to her or touched her on the shoulder, expecting him to turn up \textbf{\color{blue}{unexpectedly}} as he 'd made a habit of doing.\\
    \thickhline
    \end{tabular}
    \caption{Samples of generated literal sentences. Text in \textbf{\color{red}{bold red}}  represents the idiomatic expressions correctly included in the outputs; text in \textbf{\color{blue}{bold blue}}  represents the literal counterparts in the input sentences. text in \textbf{\color{green}{green}} represents the literal phrases that are poorly generated.}
    \label{tab:outputs2}
\end{table*}

\end{document}